\documentclass[compsoc,conference,a4paper,10pt,times]{IEEEtran}
\usepackage{amsmath,amssymb,amsfonts}
\usepackage{algorithmic}
\usepackage{graphicx}
\usepackage{textcomp}
\usepackage{bmpsize}
\usepackage{xcolor}
\usepackage{tikz}

\usepackage{booktabs}

\usepackage{enumitem}
\usepackage{threeparttable}
\usepackage{multirow}

\usepackage{subfig}

\usepackage{siunitx} 

\usepackage[
 backend=biber
,style=ieee
,minbibnames=1
,maxbibnames=5
,maxcitenames=1
,mincitenames=1
,eprint=true
,doi=true
,isbn=false
,url=true
]{biblatex}
\bibliography{references} 
\AtBeginBibliography{\footnotesize}

\DeclareSourcemap{
  \maps{
    \map{
      \step[fieldset=month, null]
      \step[fieldset=address, null]
      \step[fieldset=location, null]
      \step[fieldset=publisher, null]
      \step[fieldset=url, null]
      \step[fieldset=isbn, null]
      \step[fieldset=series, null]
      \step[fieldset=editor, null]
    }
  }
}

\usepackage{cleveref}
\Crefname{figure}{Fig.}{Figs.}

\usepackage{lipsum}

\def\BibTeX{{\rm B\kern-.05em{\sc i\kern-.025em b}\kern-.08em
    T\kern-.1667em\lower.7ex\hbox{E}\kern-.125emX}}

\usepackage{flushend}

\begin{document}

\title{Early-Stage Anomaly Detection:\\ A Study of Model Performance on Complete vs. Partial Flows}

\author{
    \IEEEauthorblockN{Adrian Pekar\IEEEauthorrefmark{1}\IEEEauthorrefmark{2}\IEEEauthorrefmark{3} and Richard Jozsa\IEEEauthorrefmark{1}}
    \IEEEauthorblockA{
        \IEEEauthorrefmark{1}
        Budapest University of Technology and Economics, M\H{u}egyetem rkp. 3., H-1111 Budapest, Hungary\\
        \IEEEauthorrefmark{2}HUN-REN-BME Information Systems Research Group, Magyar Tud\'{o}sok krt. 2, 1117 Budapest, Hungary\\
        \IEEEauthorrefmark{3}CUJO LLC, Budapest, Hungary\\
        Email: apekar@hit.bme.hu, jozsa.richard@edu.bme.hu
    }
}

\maketitle

\begin{abstract}
This study investigates the efficacy of machine learning models in network security threat detection through the critical lens of partial versus complete flow information, addressing a common gap between research settings and real-time operational needs. We systematically evaluate how a standard benchmark model, Random Forest, performs under varying training and testing conditions (complete/complete, partial/partial, complete/partial), quantifying the performance impact when dealing with the incomplete data typical in real-time environments. Our findings demonstrate a significant performance difference, with precision and recall dropping by up to 30\% under certain conditions when models trained on complete flows are tested against partial flows. 
The study also reveals that, for the evaluated dataset and model, a minimum threshold around 7 packets in the test set appears necessary for maintaining reliable detection rates, providing valuable, quantified insights for developing more realistic real-time detection strategies. 
\end{abstract}


\begin{tikzpicture}[remember picture,overlay]
\node[anchor=north, align=center, text=red, font=\small, yshift=-.6cm] at (current page.north) {The research described in this paper was accepted for presentation at the 10th International Workshop on Traffic Measurements\\ for Cybersecurity (WTMC 2025). This arXiv version includes methodological clarifications.
};
\end{tikzpicture}

\section{Introduction}

Real-time network anomaly detection is a cornerstone of modern cybersecurity, demanding systems that identify \textit{threats} (malicious activity) swiftly and accurately as network traffic unfolds. Machine learning (ML) offers powerful tools for this,
yet deploying ML detectors effectively faces a critical hurdle:
they must frequently operate on \textit{partial} or \textit{incomplete network flow information}. We define this as data captured and analyzed \textit{before} a flow naturally terminates, often due to the need for rapid decisions in monitoring systems.
Such partial flows inherently contain less data---fewer packets over shorter durations---than the \textit{complete flows} often used in research settings.

The challenge lies in bridging the gap between common research practices and operational requirements. While many influential studies in ML-based network anomaly detection develop and validate models using datasets predominantly composed of complete flows~\cite{7307098, 8543584, Kadri2024}, this approach does not fully capture the dynamics of real-time systems. It is important to acknowledge that the challenge of incomplete data is not entirely unaddressed, as we discuss later further; researchers \textit{have} explored aspects of processing incomplete flows,
focusing on topics like flow export timing and feature availability over time~\cite{Jirsik2017, Jirsik2018, 9464011}. 
However, a predominant focus in many validation studies, particularly those benchmarking algorithm performance, remains on complete flows, sometimes even when underlying packet data that could simulate partial flows is available.

This persistent reliance on complete flow data for model development creates a potential disconnect. Models trained and optimized on the rich information within complete flows may exhibit significantly different, often degraded, performance (accuracy, reliability) when deployed to make predictions using the limited data in partial flows encountered in practice. This discrepancy raises valid questions about the direct applicability and transferability of findings derived solely from complete-flow analyses to real-world security operations.

This paper addresses this gap by providing a \textit{systematic investigation into the impact of using partial versus complete network flow information during both the training and testing phases of ML model development.} Our primary contribution is \textit{quantifying the performance variations} across distinct training/testing scenarios (complete-vs-complete, partial-vs-partial, and the critical complete-vs-partial mismatch) to understand the practical implications for deploying anomaly detection systems. While prior work has examined aspects of incomplete data, our focus is on this direct, systematic comparison across training paradigms. This highlights the \textit{practical limitations and performance trade-offs} that emerge in operational environments---information crucial for security \textit{practitioners} who must deploy and trust these systems. 

To conduct this analysis, we employ Random Forest (RF), a supervised algorithm widely recognized and effective in cybersecurity~\cite{CICIDS2017, engelen2021, lanvin2023, Flood2024}. While unsupervised methods (\emph{e.g.}, Isolation Forest, Autoencoders) are common for anomaly detection, our goal is not to find the universally best detector, but rather to use a standard, well-understood supervised benchmark (RF) to specifically isolate and quantify the \textit{impact of data completeness} (partial vs. complete) on model performance, a question relevant regardless of the learning paradigm. 
Using RF provides a robust baseline to explore how a standard ML technique contends with partial flow data under controlled conditions. We experiment with three distinct scenarios: \textit{i)} models trained and tested on complete flows (baseline), \textit{ii)} models trained and tested on partial flows (ideal partial detection), and \textit{iii)} models trained on complete records but tested on partial ones (simulating a common deployment mismatch). Partial flows are generated using specific thresholds on \textit{packet counts} and \textit{flow durations}.

Our reappraisal reveals crucial insights: models trained on complete flows but tested on partial flows suffer a significant performance decline (up to 30\% drop in precision/recall under certain conditions). Conversely, models trained and tested consistently (complete/complete or partial/partial) generally maintain robustness, barring isolated cases. This directly challenges the assumption that models validated on complete flows readily generalize to real-time scenarios dominated by incomplete data. It underscores the critical need for consistency between training/testing data and deployment data, highlighting how data incompleteness can markedly impact detection capabilities.

Furthermore, our study indicates that, for the dataset examined, a minimum threshold of at least 7 packets in the test set is necessary to maintain acceptable detection rates when testing on partial flows, suggesting the need for specific model adjustments or training strategies. By focusing on aligning model evaluation with the operational realities of real-time monitoring, our research aims to guide the development of more resilient and truly effective ML-based anomaly detection systems.

The rest of this paper is organized as follows: \Cref{sec:background} presents the evaluation thresholds, dataset, and ML algorithm. \Cref{sec:methodology} details our data preprocessing and flow generation approach. \Cref{sec:results} analyzes model performance with complete versus partial flows. \Cref{sec:discussion} interprets the results and discusses limitations. \Cref{sec:RW} examines related work. \Cref{sec:conclusion} concludes the paper.

\section{Background}
\label{sec:background}

\subsection{Readiness Criteria for Flow Categorization}

In the realm of real-time anomaly detection, defining the precise moment when network flows contain sufficient information for reliable categorization poses significant challenges. In general, we distinguish between three common threshold types used to determine flow readiness for categorization:
\begin{itemize}
    \item \textit{Packet count:} establishing a minimum packet count observed in a flow.
    \item \textit{Flow size:} categorizing flows based on a specified amount of transferred data (in bytes).
    \item \textit{Time window:} defining a fixed time interval from the start of the flow for data collection.
\end{itemize}
While these threshold-based approaches each offer unique advantages, they also bring inherent challenges in real-time anomaly detection.
Different applications exhibit diverse traffic patterns; for instance, video streaming or file transfers generate large bursts of packets quickly, whereas DNS queries might involve fewer packets exchanged over a longer duration. Such variability means that flows reaching the same packet count or byte count threshold may have vastly different durations or represent different stages of communication maturity.

The time window approach, conversely, ensures a uniform time frame for decision-making but may capture insufficient data for short-lived flows or excessive, potentially redundant data for very active flows within the same window. Finding a single threshold type or value that is optimal across all types of network traffic and applications remains difficult.

A potential solution involves hybrid approaches, combining multiple criteria. However, determining the optimal balance between packet count, flow size, and duration to ensure comparable information richness across heterogeneous flows remains a complex task and an area of ongoing research.

\subsection{Selecting Thresholds for Model Evaluation}

In this paper, we focus our investigation on the performance of machine learning algorithms (MLAs) where flows are considered ready for categorization based on two well-defined and commonly considered threshold types:
\begin{itemize}
    \item First, we investigate performance under varying \textit{packet count thresholds}, focusing on the quantitative aspect of communication exchange (number of interactions).
    \item Subsequently, we explore efficacy under varying \textit{time window thresholds} (flow duration), emphasizing the temporal dynamics.
\end{itemize}
This 
approach allows us to systematically assess the behavior and effectiveness of the chosen MLA in scenarios simulating different real-time decision points based on either accumulated packet interactions or elapsed time.

By separately analyzing these two threshold mechanisms, we aim to provide a more detailed understanding of how data incompleteness, defined by these common operational constraints, affects MLA performance in real-time anomaly detection. This methodology aligns with our objective to quantify the gap between idealized complete-flow analysis and practical deployment scenarios.

\subsection{Dataset}

In this research, we utilize the CICIDS-2017 dataset~\cite{CICIDS2017}, a widely referenced collection of labeled network traffic flows designed for intrusion detection research. Generated over five days (July 3-7, 2017), the dataset includes benign traffic and a variety of simulated attacks, including DoS/DDoS, Port Scan, Brute Force, Web Attacks, Infiltration, Botnet, and Heartbleed, captured as PCAP files and pre-processed flow records. Monday's traffic is exclusively benign, while Tuesday through Friday contain mixtures of benign and malicious activities, with attack types varying by day.

This dataset has been extensively used for evaluating anomaly detection systems~\cite{CICIDS2017, Hindy2021, lanvin2023, pekar2024evaluating}, making it a suitable benchmark for comparing results and understanding how methodological choices (like using partial flows) impact outcomes. However, many prior works have relied on the provided complete flow records without explicitly investigating the performance implications when models face partial flows, which is the focus of our study. Furthermore, while CICIDS-2017 provides a valuable baseline, we acknowledge its age (data from 2017). The network traffic patterns and specific attack vectors may not fully represent the complexity of the current threat landscape, highlighting the need for future validation on more contemporary datasets. 

Due to the dataset's structure (different attacks/days) and the complexity of ensuring comparable analysis across days with differing traffic mixes, our analysis concentrates on the network traffic from \textit{Wednesday}. This day includes a mix of DoS attacks (Hulk, GoldenEye, Slowloris, Slowhttptest) aimed at exhausting server resources (availability impact) and the Heartbleed attack, which targets a specific OpenSSL vulnerability to extract sensitive memory content (confidentiality impact). The anomalies detected by our model thus correspond to these specific labeled attack types present in the Wednesday data. 

Crucially, the integrity of the original CICIDS-2017 flow records has been questioned due to identified labeling errors and inconsistencies~\cite{engelen2021, lanvin2023, Flood2024, pekar2024evaluating}. To mitigate risks to validity, our methodology does not rely on the pre-generated flows. Instead, we perform a careful regeneration and labeling of flow records directly from the original Wednesday PCAP file, following the attack descriptions provided by the dataset creators and incorporating corrections based on findings in~\cite{pekar2024evaluating}, ensuring a more accurate ground truth for our experiments.

\subsection{ML Algorithm}

In this study, we investigate the effects of incomplete flow information on network anomaly detection by employing the Random Forest algorithm. RF is an ensemble learning method based on decision trees, widely recognized for its robustness and frequent use in cybersecurity applications, including intrusion detection~\cite{CICIDS2017, engelen2021, lanvin2023, pekar2024evaluating, Flood2024}. It operates by constructing multiple decision trees during training and outputting the mode of the classes (classification) or mean prediction (regression) of the individual trees. This ensemble approach generally yields higher accuracy and better resistance to overfitting compared to individual decision trees.

Our choice of RF is motivated by several factors. First, its prevalence in the literature provides a well-established benchmark, allowing for comparison and context within existing research. Second, as a supervised learning algorithm, it allows us to directly assess classification performance (detecting known labeled anomalies) based on varying levels of input data completeness. While unsupervised methods are often used for detecting novel anomalies, RF is suitable for our specific goal: \textit{to quantify how performance on known anomaly types changes when a standard, widely-used classifier is trained or tested with partial versus complete flow information}. 

This focused methodology, centered on a single representative MLA, allows for a clear analysis of data completeness impact. While evaluating multiple algorithms is valuable future work, this approach provides a necessary baseline understanding of the challenges posed by partial flows in operational network security contexts, potentially informing the interpretation of results from other studies and guiding future work using different models.

\section{Methodology}
\label{sec:methodology}

This section details the systematic process used to generate complete and partial flow datasets from the raw packet captures and prepare them for ML analysis. 
Complementary insights, scripts, and detailed data supporting the observations and decisions described below are available as a digital artifact, accessible via~\cite{github-repo}.

\subsection{Raw Data Preprocessing}


Our methodology began with the preprocessing of the Wednesday raw packet trace file (PCAP) from the CICIDS-2017 dataset. To mitigate potential biases introduced by packet-level anomalies or inconsistencies in the capture process, we followed a strategy similar to that outlined by \citeauthor{lanvin2023}~\cite{lanvin2023}. First, we removed potential duplicate packets using the \texttt{editcap} command (\texttt{editcap -D 10000 input.pcap output.pcap}), discarding packets arriving within \SI{10000}{\micro\second} of an identical packet. Subsequently, we corrected the order of any out-of-sequence packets using the \texttt{reordercap} command (\texttt{reordercap input.pcap output.pcap}). These steps aim to create a cleaner, more consistent packet sequence as input for the flow generation process, ensuring that observed performance differences are more likely attributable to flow completeness rather than raw data artifacts.

\subsection{Flow Metering and Labeling}
\label{sec:flow_metering}

For processing the preprocessed PCAP file and generating flow records, we utilized NFStream~\cite{AOUINI2022108719}, a flexible Python-based network data analysis framework. NFStream converts raw traffic into structured flow data suitable for analysis and allows customization through its NFPlugin component. This adaptability was crucial for incorporating our specific flow labeling and generation logic directly into the metering process.

We implemented a custom labeling mechanism within NFStream. This mechanism assigns a label (\emph{e.g.}, `BENIGN', `DoS Hulk', `DoS GoldenEye') to each generated flow based on criteria derived from the original dataset documentation~\cite{CICIDS2017}, primarily using source/destination IP addresses, port numbers, and protocol associated with known attack infrastructure and time windows. This process ensures that flows are labeled according to the ground truth established for the dataset, directly addressing the malicious anomalies targeted in this study. The mechanism also handles bidirectional flows correctly. This integrated labeling approach, applied during flow generation, provides accurately labeled data tied to the specific flows generated under different timeout and export conditions.

\subsection{Preliminary Measurement}
\label{sec:initial}

To understand the characteristics of the traffic and inform the settings for generating our definitive ``complete flow'' dataset, we conducted a preliminary measurement run. This run used NFStream with idle and active timeout settings (60 and 120 seconds, respectively), reflecting potential parameters similar to the original CICIDS-2017 processing. This allowed us to observe phenomena, particularly those arising from flow segmentation caused by these timeouts, that could affect analysis if not properly handled.

\Cref{tab:anomalyBreakdown} shows the resulting breakdown of labeled flow \textit{records}. 
It separately counts records with packet payload~$>0$ and those with the payload $=$ 0. The sum of these two counts for a given attack type represents the total number of flow records generated for that attack under these preliminary timeout settings. \Cref{tab:TCP-FIN-RST-Breakdown} details the occurrence of TCP FIN/RST flags within these records.

\begin{table}[t]
\centering
\caption{Initial Labeled Flow Record Counts}
\label{tab:anomalyBreakdown}
\begin{tabular}{@{}l S[table-format=6.0] S[table-format=5.0]@{}} 
\toprule
\textbf{Flow Type} & {\textbf{Flows (payload $>0$)}} & {\textbf{Flows (payload $=0$)}} \\
\midrule
DoS GoldenEye         & 7916 & 870 \\
DoS Hulk              & 158027 & 594 \\
DoS Slowhttptest      & 3010 & 3088 \\
DoS Slowloris         & 5192 & 1805 \\
Heartbleed            & 11   & 0 \\
\midrule
\multicolumn{1}{r}{\textbf{$\sum$ Attacks}} & \bfseries 174156 & \bfseries 6357 \\
\midrule
BENIGN                & 310381 & 13580 \\
\midrule
\multicolumn{1}{r}{\textbf{$\sum$ Total}} & \bfseries 484537 & \bfseries 19937  \\ 
\bottomrule
\end{tabular}
\end{table}

\begin{table}[t]
\centering
\caption{Flow Counts with $>2$ TCP FIN or RST Flags}
\label{tab:TCP-FIN-RST-Breakdown}
\begin{tabular}{@{}l S[table-format=5.0]@{}} 
\toprule
\textbf{Metric} & {\textbf{Flow Record Count}} \\ 
\midrule
BENIGN Flows with FIN $>$ 2    & 3008 \\
Attack Flows with FIN $>$ 2    & 5819 \\
\textbf{Total Flows with FIN $>$ 2} & \bfseries 8827 \\ 
\midrule
BENIGN Flows with RST $>$ 2    & 2273 \\
Attack Flows with RST $>$ 2    & 38559 \\
\textbf{Total Flows with RST $>$ 2} & \bfseries 40832 \\ 
\bottomrule
\end{tabular}
\end{table}

Key observations from this preliminary run included:
\begin{itemize}
    \item We observed a considerable number of flows with zero packet payload (ZPL), a characteristic not associated with any of the attacks identified in the Wednesday traffic trace, as detailed in \Cref{tab:anomalyBreakdown}.
    \item A considerable number of flows featured an unusually high count of FIN and RST packets. \Cref{tab:TCP-FIN-RST-Breakdown} summarizes these counts, specifically highlighting flows with more than two FIN or RST flags. Interestingly, the attacks observed on Wednesday (various DoS attacks and Heartbleed) are not characterized by an increased number of packets with TCP FIN or RST flags. This finding suggests that the packet traces may capture not only the attack signatures but also the repercussions, such as servers potentially saturated and beginning to terminate connections by sending packets with these TCP flags.
    \item The Heartbleed attack appears as one prolonged attack that is segmented into multiple flow records by the active timeout setting.
    \item A pattern of repeated flows was identified across the dataset, distinguished using a five-tuple of source and destination IP and port numbers, alongside the protocol identifier.
    \item Numerous flows exhibited a Packet Inter-Arrival Time (PIAT) marginally below the idle timeout, set at 60 seconds. These instances might suggest that separate flows were amalgamated into a single flow due to the idle timeout configuration. Alternatively, this behavior could imply that some attacks were deliberately kept active, with packets sent just before the expiration of the timeout.
\end{itemize}

These observations highlighted the need for specific flow generation parameters and filtering steps to create a cleaner dataset focused on the core characteristics of the traffic flows, rather than measurement artifacts or secondary effects.

\subsection{Producing Complete Flows}
\label{sec:cf}

Based on the insights from the preliminary measurements (\Cref{sec:initial}), we configured NFStream to generate a refined dataset of ``complete'' flows (CF). Our goal was to capture the entirety of each relevant communication exchange while mitigating the artifacts observed previously. The following settings and filters were applied:
\begin{itemize}
\item The idle timeout is maintained at 60 seconds, 
falling within the range of typical settings for Linux Kernel netfilter and IPv4 TCP-specific networking configurations.
\item The active timeout is set to 18,000 seconds (5 hours) to prevent the segmentation of exceptionally long flows due to timeout expiration.
\item A TCP FIN/RST flag-based flow expiration policy has been implemented, which terminates flows at the first detection of either a FIN or RST flag, whether at flow initiation or upon update. This approach aims to exclude the aftermath of attacks that manifest as connection terminations. As a result, residual flow fragments, typically single-packet flows marked by a FIN or RST flag, are excluded.
\item The flow start time is now incorporated into the flow ID hash, enhancing the ability to match complete flows with their corresponding partial counterparts. Potential duplicate flow hash entries (despite this addition to the six-tuple used for unique flow identification) are discarded.
\item All flows with zero packet payloads (ZPL) are excluded.
\item The Heartbleed attack is omitted due to an insufficient number of samples for meaningful classification performance evaluation.
\end{itemize}

\begin{table}[t]
\centering
\caption{Refined Complete Flow Dataset Breakdown}
\label{tab:completeflows}
\begin{tabular}{@{}l S[table-format=6.0]@{}} 
\toprule
\textbf{Flow Type} & {\textbf{Flow Count}} \\ 
\midrule
DoS GoldenEye & 7917 \\
DoS Hulk & 158680 \\
DoS Slowhttptest & 3707 \\
DoS Slowloris & 5683 \\
\midrule
\multicolumn{1}{r}{\textbf{$\sum$ Attacks}} & \bfseries 175987 \\ 
\midrule
BENIGN & 326363 \\
\midrule
\multicolumn{1}{r}{\textbf{$\sum$ Total}} & \bfseries 502350 \\ 
\bottomrule
\end{tabular}
\end{table}


The resulting distribution of the refined CF dataset is detailed in \Cref{tab:completeflows}.
Compared to \Cref{tab:anomalyBreakdown}, it presents a marginally higher number of flow records for specific types. This is primarily due to our TCP flow expiration policy, which segments flows with multiple, irregular sequences of TCP FIN and RST flags into separate subflows. To mitigate any potential bias arising from such segmentation, post-processing steps are undertaken, where partial flows are matched with their complete counterparts using their six-tuple identification. This ensures that the initial segments of partial flows are accurately compared and aligned with the corresponding complete flows. This CF dataset represents our baseline for ``ideal'' or complete information available post-hoc.

\subsection{Producing Partial Flows}
\label{sec:pf}

To simulate real-time detection scenarios where decisions must be made before a flow completes, we generated partial flow (PF) datasets based on the CF dataset. For each complete flow in the CF dataset, we generated corresponding partial flows exported at specific thresholds using two distinct mechanisms, reflecting common real-time triggers:
\begin{enumerate} 
    \item \textit{Packet Count Threshold:} This mechanism exports a partial flow record precisely when the flow reaches a specific number of packets (from the start of the flow). We generated separate PF datasets for exact packet counts $N_{pc}$ across the range $\{2, 3, 4, \ldots, 20\}$. This simulates scenarios where analysis is triggered after a fixed number of packet exchanges.
    \item \textit{Flow Duration Threshold:} This mechanism exports a partial flow record when the flow's duration (time elapsed since the first packet) reaches a specific target time window. To account for minor timing variations, we included flows whose duration at export fell within a $\pm$20\% range of the target. We generated separate PF datasets for target durations $N_{fd}$ across the range $\{5, 10, 50, 100, 150, 300, 500, 1000, 5000, 10000,$ $15000, 20000\}$ milliseconds. This simulates analysis triggered after fixed time intervals.
\end{enumerate}

Crucially, partial flows were not labeled directly during generation. Instead, each generated partial flow retained the unique 6-tuple identifier of its corresponding complete flow from the CF dataset. This identifier was then used to assign the correct label (\emph{e.g.}, `BENIGN', `DoS Hulk') from the parent CF record to the PF record. Only partial flows that could be successfully matched to a parent CF record were retained. Furthermore, for a given PF dataset defined by a threshold (\emph{e.g.}, $N_{pc}=5$ or $N_{fd}=\SI{100}{ms}$), only the partial flows generated exactly at that threshold (or within the duration range for $N_{fd}$) were included in that specific dataset. This ensures that each PF dataset represents the information available precisely at that decision point.

This methodology allows us to directly compare the performance of ML models trained/tested on the CF dataset versus various PF datasets, systematically examining the impact of information availability defined by these common real-time export triggers (packet count and duration). This contrasts with studies relying solely on pre-defined complete flows and enables a quantitative assessment relevant to operational deployment.

\section{Results}
\label{sec:results}

This section presents the empirical evaluation designed to quantify the performance impact of using partial versus complete flow information for anomaly detection with Random Forest on the prepared CICIDS-2017 Wednesday dataset. We analyze results based on two common partial flow generation triggers: packet count and flow duration. The findings provide crucial context for interpreting previous studies that relied primarily on complete flows and highlight challenges for real-time deployment.

\textbf{Evaluation Metrics:} 
%
For \textit{binary classification} scenarios, we report precision, recall, and F1-score for the \textit{anomaly class} to focus on attack detection capability, which is the primary concern in cybersecurity applications where missing attacks has severe consequences. For \textit{multi-class} scenarios, we report \textit{macro-averaged metrics} across all classes. Due to these different evaluation focuses and the inherent class imbalance in network traffic, direct performance comparisons between binary and multi-class results should be interpreted with caution. For consistent cross-paradigm comparisons, we provide alternative metric calculations in our digital artifacts~\cite{github-repo}.

\subsection{Packet Count-based Evaluation}

We first examine scenarios where partial flows are defined by reaching a specific packet count ($N_{pc}$).

\begin{table*}[htbp]
\scriptsize
\centering
\caption{Distribution of Complete and Partial Network Flows by Packet Count Thresholds}
\label{tbl:pc_res}
\renewcommand{\arraystretch}{0.6} 
\begin{tabular}{lrrrrrrrr}
\toprule
\textbf{DS} & \textbf{TOTAL} & \textbf{BENIGN} & \textbf{ANOMALY} & \textbf{Anomaly Type} & \textbf{Count} & \textbf{Min Dur. [ms]} & \textbf{Mean Dur. [ms]} & \textbf{Max Dur. [ms]} \\
\midrule
\multirow{4}{*}{CF} & \multirow{4}{*}{502 350} & \multirow{4}{*}{326 363} & \multirow{4}{*}{175 987} & DoS GoldenEye & 7 917 & 0 & 11 028.96 & 106 793 \\
 &  &  &  & DoS Hulk & 158 680 & 0 & 693.68 & 128 843 \\
 &  &  &  & DoS Slowhttptest & 3 707 & 0 & 9 562.98 & 167 903 \\
 &  &  &  & DoS Slowloris & 5 683 & 0 & 33 548.82 & 105 745 \\
\midrule
\multirow{4}{*}{PC=2} & \multirow{4}{*}{500 493} & \multirow{4}{*}{324 508} & \multirow{4}{*}{175 985} & DoS GoldenEye & 7 917 & 0 & 130.35 & 58 333 \\
 &  &  &  & DoS Hulk & 158 680 & 0 & 456.34 & 32 096 \\
 &  &  &  & DoS Slowhttptest & 3 705 & 0 & 152.66 & 36 864 \\
 &  &  &  & DoS Slowloris & 5 683 & 0 & 168.72 & 1 000 \\
\midrule
\multirow{4}{*}{PC=3} & \multirow{4}{*}{290 145} & \multirow{4}{*}{119 229} & \multirow{4}{*}{170 916} & DoS GoldenEye & 7 568 & 0 & 204.30 & 58 333 \\
 &  &  &  & DoS Hulk & 158 559 & 0 & 528.67 & 64 257 \\
 &  &  &  & DoS Slowhttptest & 2 419 & 0 & 3 055.00 & 43 264 \\
 &  &  &  & DoS Slowloris & 2 370 & 0 & 3 403.97 & 57 769 \\
\midrule
\multirow{4}{*}{PC=4} & \multirow{4}{*}{271 751} & \multirow{4}{*}{101 923} & \multirow{4}{*}{169 828} & DoS GoldenEye & 7 567 & 0 & 233.50 & 48 079 \\
 &  &  &  & DoS Hulk & 158 557 & 0 & 529.51 & 65 721 \\
 &  &  &  & DoS Slowhttptest & 1 469 & 0 & 1 415.11 & 98 560 \\
 &  &  &  & DoS Slowloris & 2 235 & 0 & 908.58 & 3 004 \\
\midrule
\multirow{4}{*}{PC=5} & \multirow{4}{*}{263 319} & \multirow{4}{*}{93 547} & \multirow{4}{*}{169 772} & DoS GoldenEye & 7 567 & 0 & 255.96 & 48 080 \\
 &  &  &  & DoS Hulk & 158 557 & 0 & 552.16 & 65 721 \\
 &  &  &  & DoS Slowhttptest & 1 413 & 1 & 2 068.20 & 15 036 \\
 &  &  &  & DoS Slowloris & 2 235 & 0 & 989.58 & 3 005 \\
\midrule
\multirow{4}{*}{PC=6} & \multirow{4}{*}{261 477} & \multirow{4}{*}{92 476} & \multirow{4}{*}{169 001} & DoS GoldenEye & 7 567 & 0 & 4 773.53 & 52 865 \\
 &  &  &  & DoS Hulk & 158 557 & 0 & 637.06 & 66 719 \\
 &  &  &  & DoS Slowhttptest & 679 & 8 & 6 174.75 & 31 079 \\
 &  &  &  & DoS Slowloris & 2 198 & 1 & 2 466.67 & 53 080 \\
\midrule
\multirow{4}{*}{PC=7} & \multirow{4}{*}{260 091} & \multirow{4}{*}{91 375} & \multirow{4}{*}{168 716} & DoS GoldenEye & 7 567 & 0 & 5 415.78 & 63 095 \\
 &  &  &  & DoS Hulk & 158 461 & 0 & 659.39 & 66 719 \\
 &  &  &  & DoS Slowhttptest & 631 & 9 & 8 649.36 & 63 160 \\
 &  &  &  & DoS Slowloris & 2 057 & 2 & 3 017.98 & 54 078 \\
\midrule
\multirow{4}{*}{PC=8} & \multirow{4}{*}{251 718} & \multirow{4}{*}{87 891} & \multirow{4}{*}{163 827} & DoS GoldenEye & 7 567 & 1 & 5 743.47 & 63 095 \\
 &  &  &  & DoS Hulk & 153 941 & 0 & 688.26 & 68 723 \\
 &  &  &  & DoS Slowhttptest & 458 & 11 & 11 213.53 & 63 160 \\
 &  &  &  & DoS Slowloris & 1 861 & 3 & 2 364.51 & 54 079 \\
\midrule
\multirow{4}{*}{PC=9} & \multirow{4}{*}{222 951} & \multirow{4}{*}{85 492} & \multirow{4}{*}{137 459} & DoS GoldenEye & 7 411 & 1 & 6 757.70 & 63 095 \\
 &  &  &  & DoS Hulk & 127 790 & 0 & 784.01 & 68 723 \\
 &  &  &  & DoS Slowhttptest & 415 & 17 & 12 643.28 & 63 161 \\
 &  &  &  & DoS Slowloris & 1 843 & 3 & 3 583.45 & 59 080 \\
\midrule
\multirow{4}{*}{PC=10} & \multirow{4}{*}{141 273} & \multirow{4}{*}{84 555} & \multirow{4}{*}{56 718} & DoS GoldenEye & 6 183 & 1 & 6 603.48 & 63 096 \\
 &  &  &  & DoS Hulk & 48 451 & 1 & 1 602.42 & 72 731 \\
 &  &  &  & DoS Slowhttptest & 246 & 2 603 & 20 056.63 & 63 161 \\
 &  &  &  & DoS Slowloris & 1 838 & 21 & 5 688.29 & 59 080 \\
\midrule
\multirow{4}{*}{PC=11} & \multirow{4}{*}{102 837} & \multirow{4}{*}{82 489} & \multirow{4}{*}{20 348} & DoS GoldenEye & 5 585 & 2 & 6 550.69 & 63 096 \\
 &  &  &  & DoS Hulk & 12 706 & 2 & 2 787.30 & 72 731 \\
 &  &  &  & DoS Slowhttptest & 239 & 2 655 & 22 290.57 & 63 364 \\
 &  &  &  & DoS Slowloris & 1 818 & 22 & 9 709.68 & 50 039 \\
\midrule
\multirow{4}{*}{PC=12} & \multirow{4}{*}{91 649} & \multirow{4}{*}{80 104} & \multirow{4}{*}{11 545} & DoS GoldenEye & 5 253 & 2 & 7 293.73 & 68 586 \\
 &  &  &  & DoS Hulk & 4 245 & 10 & 2 991.83 & 80 753 \\
 &  &  &  & DoS Slowhttptest & 236 & 7 619 & 25 932.32 & 63 568 \\
 &  &  &  & DoS Slowloris & 1 811 & 22 & 18 236.57 & 25 680 \\
\midrule
\multirow{4}{*}{PC=13} & \multirow{4}{*}{83 175} & \multirow{4}{*}{75 062} & \multirow{4}{*}{8 113} & DoS GoldenEye & 4 366 & 2 & 8 688.62 & 68 586 \\
 &  &  &  & DoS Hulk & 1 716 & 10 & 4 343.46 & 80 754 \\
 &  &  &  & DoS Slowhttptest & 234 & 13 088 & 29 969.69 & 63 976 \\
 &  &  &  & DoS Slowloris & 1 797 & 22 & 25 719.71 & 31 622 \\
\midrule
\multirow{4}{*}{PC=14} & \multirow{4}{*}{75 628} & \multirow{4}{*}{70 410} & \multirow{4}{*}{5 218} & DoS GoldenEye & 2 560 & 3 & 9 709.52 & 68 586 \\
 &  &  &  & DoS Hulk & 642 & 161 & 8 026.42 & 96 779 \\
 &  &  &  & DoS Slowhttptest & 234 & 13 088 & 33 832.93 & 64 792 \\
 &  &  &  & DoS Slowloris & 1 782 & 212 & 30 787.90 & 32 624 \\
\midrule
\multirow{4}{*}{PC=15} & \multirow{4}{*}{69 081} & \multirow{4}{*}{65 645} & \multirow{4}{*}{3 436} & DoS GoldenEye & 1 246 & 5 & 11 117.28 & 68 587 \\
 &  &  &  & DoS Hulk & 181 & 161 & 18 406.82 & 96 779 \\
 &  &  &  & DoS Slowhttptest & 234 & 13 089 & 41 071.45 & 66 428 \\
 &  &  &  & DoS Slowloris & 1 775 & 222 & 42 131.41 & 51 365 \\
\midrule
\multirow{4}{*}{PC=16} & \multirow{4}{*}{64 929} & \multirow{4}{*}{62 137} & \multirow{4}{*}{2 792} & DoS GoldenEye & 743 & 10 & 13 591.32 & 106 793 \\
 &  &  &  & DoS Hulk & 61 & 1 612 & 57 281.75 & 128 842 \\
 &  &  &  & DoS Slowhttptest & 234 & 13 110 & 55 096.79 & 107 103 \\
 &  &  &  & DoS Slowloris & 1 754 & 222 & 70 441.35 & 102 735 \\
\midrule
\multirow{4}{*}{PC=17} & \multirow{4}{*}{61 237} & \multirow{4}{*}{58 779} & \multirow{4}{*}{2 458} & DoS GoldenEye & 487 & 12 & 16 609.48 & 106 793 \\
 &  &  &  & DoS Hulk & 43 & 1 612 & 77 516.74 & 128 843 \\
 &  &  &  & DoS Slowhttptest & 234 & 13 110 & 67 436.24 & 107 773 \\
 &  &  &  & DoS Slowloris & 1 694 & 235 & 87 924.51 & 103 735 \\
\midrule
\multirow{4}{*}{PC=18} & \multirow{4}{*}{57 217} & \multirow{4}{*}{55 799} & \multirow{4}{*}{1 418} & DoS GoldenEye & 313 & 1 661 & 18 334.05 & 105 645 \\
 &  &  &  & DoS Hulk & 13 & 1 650 & 14 255.54 & 15 483 \\
 &  &  &  & DoS Slowhttptest & 202 & 13 150 & 74 849.35 & 107 773 \\
 &  &  &  & DoS Slowloris & 890 & 272 & 104 104.38 & 105 745 \\
\midrule
\multirow{4}{*}{PC=19} & \multirow{4}{*}{54 154} & \multirow{4}{*}{53 267} & \multirow{4}{*}{887} & DoS GoldenEye & 206 & 2 746 & 21 941.75 & 105 645 \\
 &  &  &  & DoS Hulk & 13 & 2 051 & 14 449.23 & 16 355 \\
 &  &  &  & DoS Slowhttptest & 174 & 18 112 & 85 884.79 & 111 750 \\
 &  &  &  & DoS Slowloris & 494 & 1 675 & 105 045.52 & 105 745 \\
\midrule
\multirow{4}{*}{PC=20} & \multirow{4}{*}{51 267} & \multirow{4}{*}{50 961} & \multirow{4}{*}{306} & DoS GoldenEye & 150 & 2 746 & 24 907.24 & 73 588 \\
 &  &  &  & DoS Hulk & 13 & 2 055 & 24 756.85 & 27 741 \\
 &  &  &  & DoS Slowhttptest & 141 & 52 390 & 100 305.45 & 119 800 \\
 &  &  &  & DoS Slowloris & 2 & 2 250 & 6 966.50 & 11 683 \\

\bottomrule
\end{tabular}
\end{table*}

\begin{figure*}[t]
    \centering 
    \subfloat[Binary classification performance across packet count scenarios.]{%
        \includegraphics[width=0.5\textwidth]{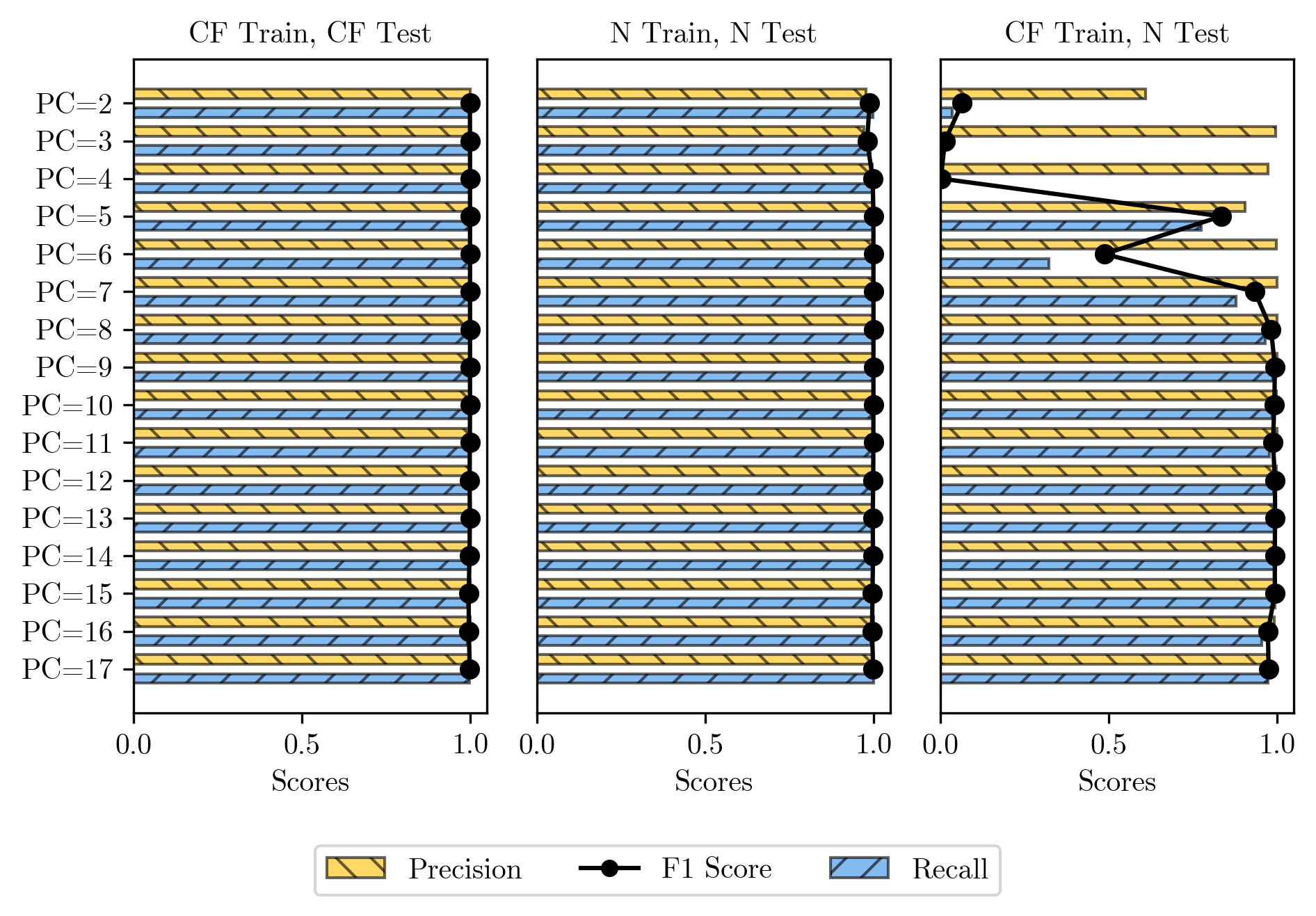}
        \label{fig:pc_b}%
    }
    \subfloat[Multi-class classification performance across packet count scenarios.]{%
        \includegraphics[width=0.5\textwidth]{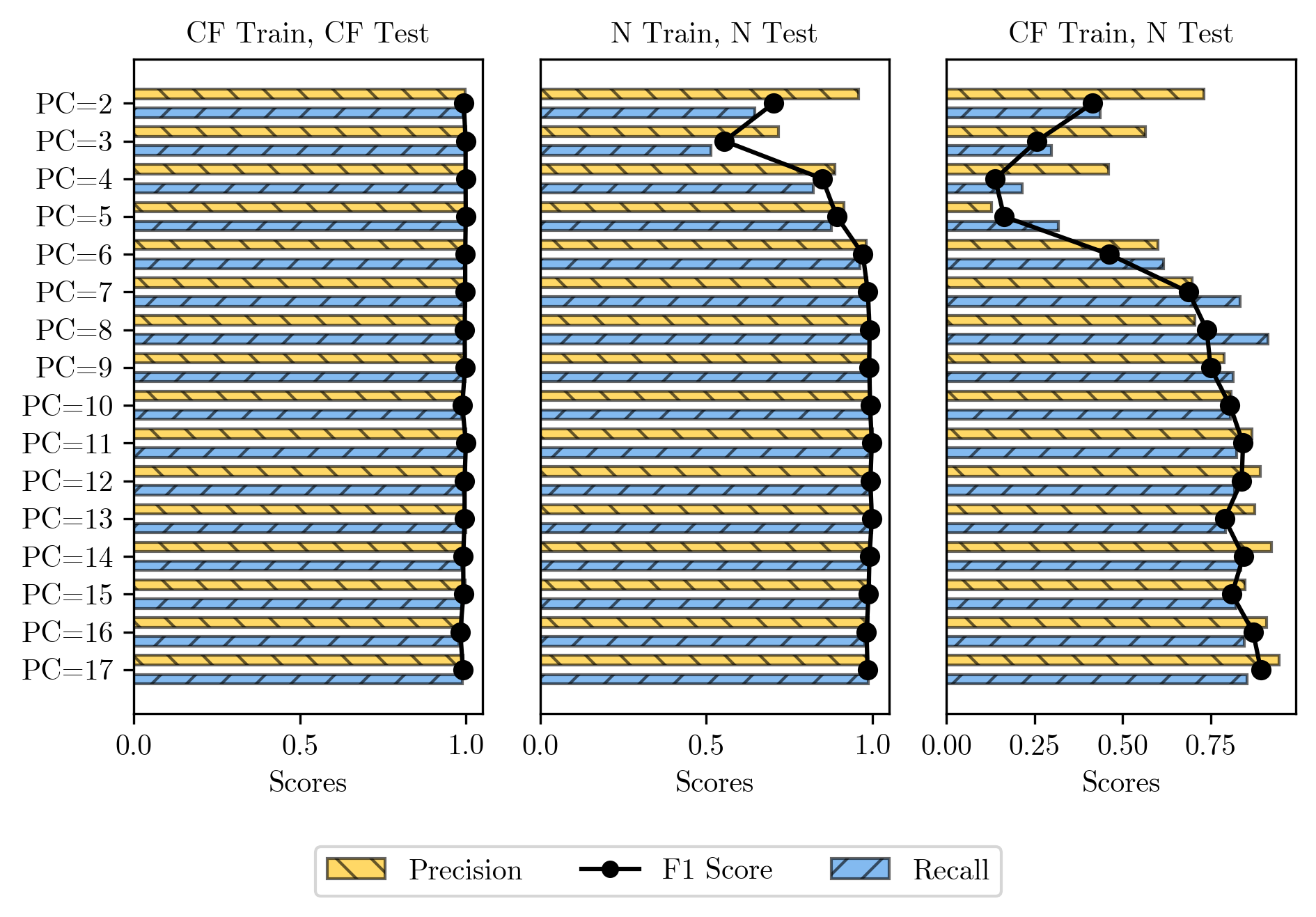}
        \label{fig:pc_m}%
    }
    \caption{Comparative analysis of binary and multi-class classification performance across different packet counts.}
    \label{fig:pc}
\end{figure*}

\subsubsection{CF and PF Dataset Distribution}

\Cref{tbl:pc_res} presents the dataset distributions for packet count-based evaluation. The table starts with the Complete Flows (CF) category, which encompasses the ground truth dataset (see \Cref{sec:cf}) and then breaks down into partial flows labeled as PC=N, where N ranges from 2 to 20, representing datasets characterized by specific packet counts. For each category, \Cref{tbl:pc_res} lists the total number of flows, classified into benign and anomalous types, with anomalies further subdivided into distinct types of DoS attacks. The table also includes metrics on the minimum, mean, and maximum durations (in milliseconds) for these categories.

Analysis of \Cref{tbl:pc_res} reveals a significant decrease in the total number of flows as the packet count per flow increases. The dataset starts with 502,350 network flows in the CF category, predominantly benign (326,363), with 175,987 flows categorized as anomalous, spanning four types of DoS attacks. In contrast, at PC=20, the dataset contains only 51,267 flows, with 50,961 being benign and merely 306 classified as anomalous.

This trend continues across specific attack types: as packet counts in flows increase, the number of records correspondingly decreases, suggesting that many attacks transmit only a few packets. Moreover, there is an observable increase in the maximum duration of attacks as packet count per flow increases, especially notable in Slowhttptest and Slowloris attacks, which are known for prolonged durations. Thus, the packet count per flow crucially influences the observed characteristics of network attacks, with higher packet counts typically associated with longer durations. Conversely, many attacks consist of flows with few packets, indicating that extracted flow statistics might lack sufficient information for precise anomaly detection, complicating effective detection mechanisms.

\subsubsection{CF vs. PF Performance Comparison}
\label{sec:results_pc_perf}

\Cref{fig:pc} illustrates the precision, F\textsubscript{1}-score, and recall for both binary (\Cref{fig:pc_b}) and multi-class (\Cref{fig:pc_m}) classification under three scenarios: 
\begin{itemize}
    \item CF used for both training and testing, 
    \item PC=N for both training and testing, and 
    \item CF used for training with PC=N for testing.
\end{itemize}

Note that PC ranges from 2 to 17, unlike in \Cref{tbl:pc_res}, because PCs 18, 19, and 20 lacked enough samples for meaningful comparison. Additionally, before training, we selected only those flows from the CF dataset that intersect with the flow hash used as identification in the respective partial flow dataset.

From \Cref{fig:pc_b}, it is observed that both CF train/CF test and N train/N test scenarios achieve sufficient performance starting at PC=2, though performance is slightly lower at minimal PC values in the N train/N test scenario. However, the CF train/N test scenario highlights the negative impacts of insufficient information due to lower packet counts, indicating that at least 8 packets are necessary in the N test to achieve usable performance using CF train model, with some minor fluctuations at higher packet counts.

The multi-class classification in \Cref{fig:pc_m} similarly shows that low packet counts negatively affect the N train/N test scenario, requiring at least 6 packets to achieve acceptable performance. Additionally, even the highest packet count seems inadequate for stable performance across precision, F\textsubscript{1}-score, and recall in the CF train/N test scenario, underscoring the challenges of anomaly detection with limited data when only completed flows have been used for training.

The results show that conclusions drawn from evaluations using only complete flows may not generalize to operational environments. Notably, while previous works have reported high detection rates using this dataset, our findings suggest that such performance metrics should be interpreted with caution when considering real-time detection scenarios.

\subsection{Flow Duration-based Evaluation}

Next, we analyze scenarios where partial flows are defined by reaching a specific flow duration ($N_{fd}$) threshold.

\begin{table*}[htbp]
\scriptsize
\centering
\caption{Distribution of Complete and Partial Network Flows by Flow Duration Thresholds}
\label{tbl:fd_res}
\renewcommand{\arraystretch}{0.6} 
\begin{tabular}{lrrrrrrrr}
\toprule
\textbf{DS} & \textbf{TOTAL} & \textbf{BENIGN} & \textbf{ANOMALY} & \textbf{Anomaly Type} & \textbf{Count} & \textbf{Min Pckts.} & \textbf{Mean Pckts.} & \textbf{Max Pckts.} \\
\midrule
\multirow{4}{*}{CF} & \multirow{4}{*}{502 350} & \multirow{4}{*}{326 363} & \multirow{4}{*}{175 987} & DoS GoldenEye & 7 917 & 2 & 12.14 & 30 \\
 & & & & DoS Hulk & 158 680 & 2 & 9.20 & 27 \\
 & & & & DoS Slowhttptest & 3 707 & 1 & 4.72 & 35 \\
 & & & & DoS Slowloris & 5 683 & 2 & 7.36 & 27 \\
\midrule

\multirow{3}{*}{FD=5ms} & \multirow{3}{*}{7 693} & \multirow{3}{*}{729} & \multirow{3}{*}{6 964} & DoS GoldenEye & 270 & 3 & 7.11 & 15 \\
 & & & & DoS Hulk & 5 917 & 3 & 4.03 & 10 \\
 & & & & DoS Slowhttptest & 777 & 2 & 3.98 & 4 \\
\midrule

\multirow{2}{*}{FD=10ms} & \multirow{2}{*}{82 521} & \multirow{2}{*}{1 093} & \multirow{2}{*}{81 428} & DoS GoldenEye & 107 & 3 & 9.22 & 14 \\
 & & & & DoS Hulk & 81 321 & 2 & 3.16 & 11 \\
\midrule

FD=50ms & 53 984 & 48 309 & 5 675 & DoS Hulk & 5 675 & 3 & 7.37 & 11 \\
\midrule

FD=100ms & 46 776 & 42 783 & 3 993 & DoS Hulk & 3 993 & 4 & 7.27 & 11 \\
\midrule

\multirow{2}{*}{FD=150ms} & \multirow{2}{*}{104 191} & \multirow{2}{*}{37 334} & \multirow{2}{*}{66 857} & DoS Hulk & 66 738 & 6 & 7.81 & 11 \\
 & & & & DoS Slowhttptest & 119 & 2 & 2.44 & 4 \\
\midrule

FD=300ms & 31 784 & 25 597 & 6 187 & DoS Hulk & 6 187 & 5 & 7.16 & 12 \\
\midrule

\multirow{2}{*}{FD=500ms} & \multirow{2}{*}{20 237} & \multirow{2}{*}{17 266} & \multirow{2}{*}{2 971} & DoS Hulk & 2 832 & 8 & 8.18 & 13 \\
 & & & & DoS Slowloris & 139 & 5 & 6.02 & 7 \\
\midrule

\multirow{4}{*}{FD=1000ms} & \multirow{4}{*}{24 545} & \multirow{4}{*}{9 606} & \multirow{4}{*}{14 939} & DoS GoldenEye & 442 & 2 & 4.08 & 8 \\
 & & & & DoS Hulk & 13 856 & 2 & 5.56 & 14 \\
 & & & & DoS Slowhttptest & 152 & 2 & 3.16 & 8 \\
 & & & & DoS Slowloris & 489 & 2 & 4.22 & 7 \\
\midrule

FD=5000ms & 15 953 & 13 714 & 2 239 & DoS GoldenEye & 2 239 & 6 & 8.37 & 18 \\
\midrule

FD=10000ms & 37 060 & 31 689 & 5 371 & DoS GoldenEye & 5 371 & 6 & 12.10 & 24 \\
\midrule

\multirow{4}{*}{FD=15000ms} & \multirow{4}{*}{8 763} & \multirow{4}{*}{4 243} & \multirow{4}{*}{4 520} & DoS GoldenEye & 510 & 2 & 13.11 & 27 \\
 & & & & DoS Hulk & 3 314 & 2 & 2.06 & 19 \\
 & & & & DoS Slowhttptest & 138 & 5 & 6.43 & 13 \\
 & & & & DoS Slowloris & 558 & 6 & 12.68 & 25 \\
\midrule

\multirow{2}{*}{FD=20000ms} & \multirow{2}{*}{30 545} & \multirow{2}{*}{29 864} & \multirow{2}{*}{681} & DoS GoldenEye & 341 & 8 & 14.11 & 24 \\
 & & & & DoS Slowhttptest & 340 & 3 & 5.01 & 19 \\

\bottomrule
\end{tabular}
\end{table*}

\begin{figure*}[t]
    \centering 
    \subfloat[Binary classification performance across flow duration scenarios.]{%
        \includegraphics[width=0.46\textwidth]{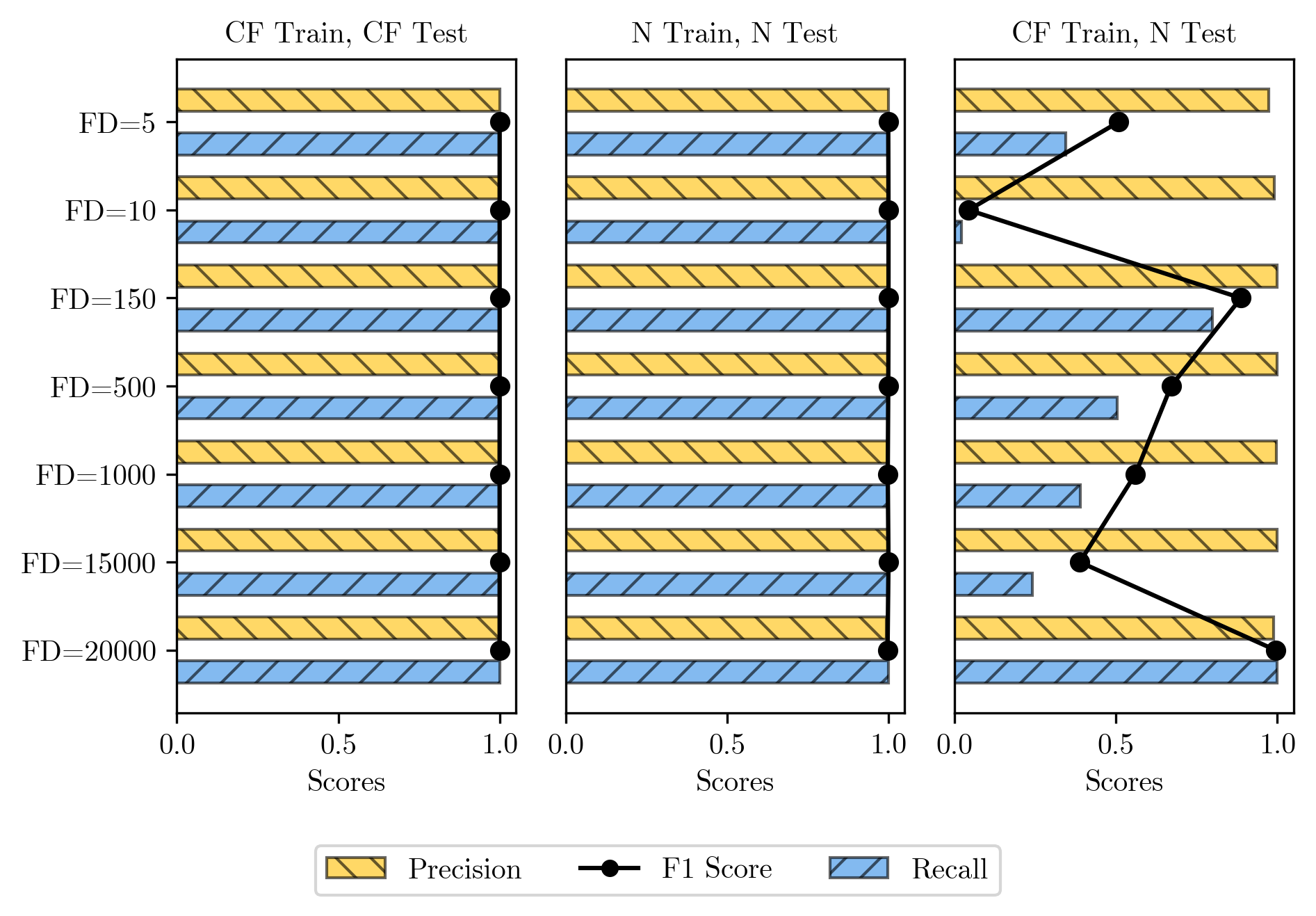}
        \label{fig:fd_b}%
    }
    \subfloat[Multi-class classification performance across flow duration scenarios.]{%
        \includegraphics[width=0.46\textwidth]{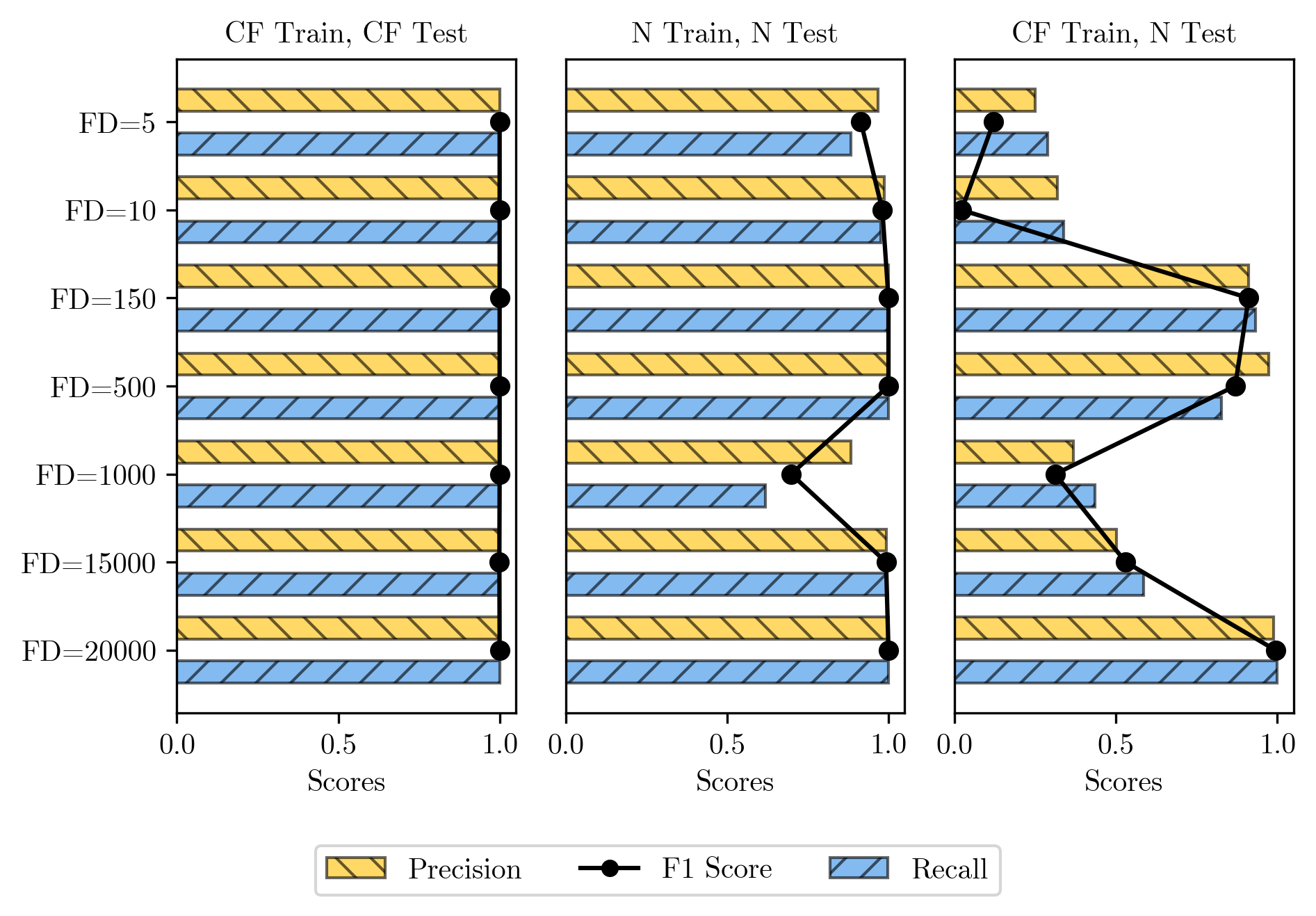}
        \label{fig:fd_m}%
    }

    \caption{Comparative analysis of binary and multi-class classification performance across different flow durations}. 
    \label{fig:fd}
\end{figure*}

\subsubsection{CF and PF Dataset Distribution}

Similar to \Cref{tbl:pc_res}, \Cref{tbl:fd_res} outlines the dataset distributions for flow duration-based evaluation. This methodical categorization delineates a transition from Complete Flows (CF) to progressively more granular Partial Flows (PF), identified as FD=N, where N encompasses a range of specified durations $5, 10, 50, 100, 150, 300, 500, 1000, 5000, 10000, 15000,$ $20000$ milliseconds. Each category within \Cref{tbl:fd_res} specifies the total number of flows, distinguishing between benign and anomalous flows, with the latter further divided into distinct DoS attack types. Contrasting with \Cref{tbl:pc_res}, which detailed flow durations, this table provides insights on the minimum, mean, and maximum packet counts for these categories.

Upon examining \Cref{tbl:fd_res}, a similar trend emerges as was seen with packet count-based datasets; however, when flow duration thresholds define partial flow generation, there is a dramatic drop in flow counts for each dataset. This decrease persists even though our methodology accommodates flows with durations within $\pm$20\% of the threshold values, as detailed in \Cref{sec:pf}.

The statistics in flow count with various flow durations underscores the challenge of capturing meaningful data within constricted time frames, particularly for anomaly detection where comprehensive flow information is crucial. Across all flow we see a fluctuating count of flows per most cateogries. For example, for DoS Hulk with FD=10ms, there are 81,321 flows, which decreases to 66,738 flows at FD=150ms, drops further to 2,832 at FD=500ms, and then unexpectedly rises again to 13,856 at FD=1000ms. Contrary to expectations, the flow count does not monotonically increase with longer durations.

The irregular trend could result from the inherent nature of network traffic during a DoS Hulk attack, where periods of intense activity are interspersed with lulls. As the duration threshold increases, the dataset may initially miss capturing these shorter bursts of activity, resulting in a lower count. Yet as the threshold extends further, it may again encompass subsequent waves of attack traffic, thus accounting for the rise in flow count.

\subsubsection{CF vs. PF Performance Comparison}
\label{sec:results_fd_perf}

\Cref{fig:fd} depicts the performance metrics of precision, F\textsubscript{1}-score, and recall for binary and multi-class classifications across a range of FD thresholds. For the sake of this analysis, only FD=5, 10, 150, 500, 1000, 15000, and 20000 ms are included, as other durations did not have enough data to yield statistically significant results. Analogous to the packet count-based evaluation, prior to training, we exclusively selected flows from the CF dataset that correspond with the flow hash identifiers present within the respective partial flow dataset.

\Cref{fig:fd_b} confirms the trend observed in packet count evaluations. When the training and testing datasets match (CF train/CF test and N train/N test), binary classification performance is strong and stable. However, the model's ability to accurately classify benign and anomalous flows is compromised in the CF train/N test scenario, reflecting the challenge of applying a model trained on complete flow data to partial flows characterized by various durations.

In \Cref{fig:fd_m}, this issue is further exacerbated in the multi-class classification scenario. Here, even the N train/N test performance demonstrates a notable decline, particularly at FD=1000ms, suggesting that as with certain flow durations set as threshold, the model's ability to distinguish between multiple classes of traffic—beyond benign and anomalous—becomes is unreliable. This decline in performance may be attributed to the reduced information content in flows with various duration, which does not adequately capture the complexities of multi-class traffic patterns.

The performance dip in the CF train/N test scenario for multi-class classification is even more pronounced, reinforcing the notion that comprehensive flow data is crucial for developing robust models. The precision, F\textsubscript{1}-score, and recall all experience a steep drop when the model trained on complete data is tested on partial data. This emphasizes the model's dependence on the quantity and quality of training data and the challenge of applying it to substantially different test conditions.

The irregular trends observed in flow counts and performance metrics highlight a critical aspect often overlooked in previous evaluations: the complex relationship between flow duration and detection capability. This relationship has significant implications for how we interpret past studies that assumed complete flow information would be available.



\section{Discussion}
\label{sec:discussion}

Our empirical results provide several critical insights into the performance of ML-based network anomaly detection when faced with the reality of partial flow information. This discussion synthesizes these findings, considers their implications, and outlines limitations and future research avenues.

\subsection{Implications of Findings}

The systematic comparison across complete and partial flow scenarios reveals significant implications.
%
    Our primary finding is the quantifiable performance degradation (up to 30\% drop in precision/recall in some scenarios) when models trained on CF are tested on PF, particularly those defined by low packet counts or short durations (\Cref{fig:pc}, \Cref{fig:fd}). This directly challenges the implicit assumption in many studies that high performance achieved using CF translates directly to real-time operational effectiveness. Practitioners should be cautious when interpreting lab-based results derived solely from CF.
    
    Next, the study highlights differing sensitivities to threshold types. While consistent performance was achievable in packet count scenarios (PC=N Train/PC=N Test) above a certain threshold (around 7 packets for the CF Train/PF Test binary case), performance based on flow duration thresholds was more volatile (\Cref{tbl:fd_res}, \Cref{fig:fd}). This suggests packet count might offer a more stable trigger for early analysis than fixed time windows for this dataset and model, although both require careful consideration.
    
    The observation that approximately 7-8 packets were needed in the test set for a CF-trained RF model to achieve reasonable binary classification performance (\Cref{sec:results_pc_perf}) provides a concrete, albeit dataset-specific, benchmark for the minimum information required under these mismatched conditions. This finding has direct implications for designing early detection systems.
    
    The results also underscore the inherent trade-off between detection speed (requiring early decisions on partial data) and accuracy/reliability (which often improves with more complete data). The poor performance in the CF Train/PF Test multi-class scenarios further emphasizes that complex discriminations require richer information, which may not be available early in a flow's lifecycle.
    
    Lastly, the common practice of evaluating NIDS primarily on complete flow datasets may provide an overly optimistic view of real-world capability. Our work advocates for evaluation methodologies that explicitly incorporate partial flow scenarios, particularly the CF Train/PF Test mismatch, to better assess practical deployability.

\subsection{Limitations and Future Directions}

While this study provides valuable quantitative insights, certain limitations define avenues for future research.

    \textit{Evasion Potential:} The finding that around 7 packets were needed for reliable detection by the CF-trained model (\Cref{sec:results_pc_perf}) highlights a potential vulnerability. Sophisticated attackers could potentially evade such early detection by mimicking benign behavior for the initial packets before launching malicious activity. Addressing this requires further research into adaptive detection mechanisms, analysis of feature evolution within flows over time, robust feature engineering less susceptible to mimicry, and potentially combining early detection with later-stage analysis as part of a defense-in-depth strategy. This vulnerability underscores that reliance solely on very early detection can be risky. 
    
    \textit{Generalizability---Dataset:} Our analysis is based on a single dataset (CICIDS-2017, Wednesday traffic). While chosen for its prevalence and availability of packet data, its age (2017 traffic) and known issues~\cite{engelen2021, lanvin2023, Flood2024, pekar2024evaluating} mean the findings require validation on newer, more diverse datasets reflecting contemporary traffic patterns, attack vectors, and network conditions (\emph{e.g.}, varying baseline traffic, different protocols). Real-world network traces would provide the ultimate test. 
    
    \textit{Generalizability---Algorithm:} We focused on Random Forest as a representative and widely used algorithm. While this provided a clear baseline, the extent to which these specific quantitative results (\emph{e.g.}, the 7-packet threshold) generalize to other ML algorithms (\emph{e.g.}, SVM, Deep Learning models like LSTMs or CNNs, or unsupervised methods like Isolation Forest or Autoencoders) is an open question. Future work should compare different algorithmic approaches under the same partial flow constraints. 
    
    \textit{Study Assumptions Rationale:} Examining the CF Train / PF Test scenario was deliberate. It directly models the common situation where research prototypes (often CF-trained) are considered for deployment in environments dealing with PFs. Quantifying this gap is crucial for practitioners. While training specialized models for different packet counts/durations might seem ideal, the unpredictable nature of real-world flows makes maintaining numerous specialized models challenging. Our study informs the design of single, more robust models tolerant to varying completeness. Hybrid approaches or adaptive models remain interesting future directions. 
    
    \textit{Variable-length Flows \& Feature Evolution:} Our fixed-threshold approach (PC=N, FD=N) simplifies analysis but does not fully capture the dynamic evolution of features within a single flow over time. Future work could explore methods that analyze sequences of partial flow updates or use techniques sensitive to feature changes as a flow progresses.
    
    \textit{Protocol Complexities:} Modern encrypted protocols (TLS 1.3, QUIC) often require more packets for handshake completion than our lower packet count thresholds (\emph{e.g.}, PC=2). Analyzing the impact of partial information specifically within the context of these complex handshakes and application data phases is needed.
    
    \textit{Hardware Offloading Impact:} Hardware offloading can reduce the visibility of flows after an initial packet count, posing challenges for continuous monitoring. Investigating detection techniques robust to such hardware limitations is warranted.
    
    \textit{Flow Duration Methodology Refinement:} The volatility observed in datasets generated by duration thresholds, even with flexibility, suggests that more sophisticated time-based sampling or feature extraction methods might be needed for robust duration-based early detection.

These limitations highlight opportunities for refinement and extension, building upon the foundation established here regarding the critical role of data completeness in anomaly detection performance.

\section{Related Work}
\label{sec:RW}

Our work intersects with research on early-stage network traffic analysis and studies specifically evaluating NIDS performance under varying conditions.

\subsection{Early-Stage Network Traffic Analysis}

The challenge of making timely decisions based on incomplete network data has been explored previously. Many approaches utilize sliding time windows to aggregate statistics for prediction. For example, \citeauthor{8855360}~\cite{8855360} used LSTMs on statistics from fixed time windows (\(\Delta T = 1s\)), varying the window count (\(W\)). Others focus on the first few packets; \citeauthor{8990084}~\cite{8990084} applied CNNs and Autoencoders to the first \(N\) packets directly, while \citeauthor{9269112}~\cite{9269112} used Autoencoders with varying time windows (10s, 10ms) for DDoS/Port Scan detection. Techniques like threshold-based nearest neighbors combined with Autoencoders~\cite{8260756} and periodic monitoring with varying micro-window durations~\cite{7740019} have also been proposed. Autoencoders on direction-dependent flow statistics were used by~\citeauthor{9110372}~\cite{9110372} for DDoS detection.

While valuable, these studies often focus on a single mechanism (\emph{e.g.}, time windows, fixed packet count) or algorithm. Our work provides a direct, systematic comparison of performance across three distinct train/test paradigms (CF/CF, PF/PF, CF/PF) using \textit{both} packet count and flow duration as triggers for partial flow generation. The key contribution is the quantification of the performance impact specifically due to the train/test mismatch using these common triggers, providing a nuanced understanding relevant to practical deployment trade-offs.

\subsection{Comparison with Relevant Research}

A particularly relevant recent study by \citeauthor{giryes2024flow}~\cite{giryes2024flow} proposed a Set-Tree model (compatible with RF) for intrusion detection, processing flows as streams of packet headers. Using CICIDS-2017, they trained on complete streams but tested using only the first few packets, reporting high accuracy with just 2 or 4 packets often being sufficient.

This contrasts sharply with our finding that roughly 7 packets were necessary for reasonable performance in our CF Train / PF Test binary scenario using RF. We hypothesize this discrepancy stems from several factors. Firstly, the known methodological flaws and potential labeling errors within the CICIDS-2017 dataset~\cite{lanvin2023, engelen2021, Flood2024, pekar2024evaluating} could lead to models overfitting to artifacts easily distinguishable within the first few packets, especially if evaluation does not use carefully corrected ground truth or considers the dataset's limitations. The claim by \citeauthor{giryes2024flow} of high performance using only the first 4 packets (where typically only 1 packet carries application data after a TCP handshake in this dataset) without retraining warrants scrutiny regarding potential overfitting or reliance on dataset-specific artifacts. Secondly, our methodology explicitly simulated different train/test conditions, quantifying the drop, whereas their focus was on the capability of their specific model architecture when tested on prefixes. Thirdly, the increasing prevalence of protocols like TLS, requiring more handshake packets before application data exchange, further questions the real-world applicability of relying on only 2-4 packets for general intrusion detection. Our study, by systematically evaluating performance across different packet counts and highlighting the ${\sim}7$ packet requirement under mismatch conditions, emphasizes the need for robust evaluation frameworks considering data completeness and potential dataset issues for realistic assessments.

\section{Conclusion}
\label{sec:conclusion}

This study presents a systematic reappraisal of how partial flow information affects machine learning-based anomaly detection. Using a refined version of the CICIDS-2017 dataset, we demonstrate that at least 7 packets are necessary for reliable detection, contrasting with recent studies suggesting fewer packets may suffice. Our findings have important implications for how we interpret previous evaluations using this dataset, particularly regarding the gap between laboratory results and operational requirements.
The methodology and insights presented here contribute to better evaluation practices for network security tools. By explicitly considering the impact of partial flows, we highlight a critical aspect often overlooked in previous studies. This work also demonstrates the importance of carefully examining assumptions about data completeness in security evaluations.



\section{Reproducibility and Extended Analysis} 

To support reproducibility and facilitate further investigation of our findings, a digital artifact containing all necessary data and code is publicly available at~\cite{github-repo}. This includes the carefully processed datasets used for training and testing (aligned with the methodologies in \Cref{sec:methodology}), the analysis scripts used to generate the results, and performance metrics. Beyond the core metrics 
presented in \Cref{sec:results}, the artifact also provides extended analyses, 
and complementary visualizations, offering deeper insights into specific class performance and error types.

\section*{Acknowledgement}

Supported by the János Bolyai Research Scholarship of the Hungarian Academy of Sciences. 
%
This work was also supported by projects TKP2021-NVA-02 (financed under the TKP2021-NVA scheme) and 2024-1.2.6-EUREKA-2024-00009 (financed under the 2024-1.2.6-EUREKA scheme), implemented with support provided by the Ministry of Culture and Innovation of Hungary from the National Research, Development and Innovation Fund.

\printbibliography

\end{document}